\def\year{2021}\relax
\documentclass[letterpaper]{article} % DO NOT CHANGE THIS
\usepackage{aaai21}  % DO NOT CHANGE THIS
\usepackage{times}  % DO NOT CHANGE THIS
\usepackage{helvet} % DO NOT CHANGE THIS
\usepackage{courier}  % DO NOT CHANGE THIS
\usepackage[hyphens]{url}  % DO NOT CHANGE THIS
\usepackage{graphicx} % DO NOT CHANGE THIS
\usepackage{natbib}  % DO NOT CHANGE THIS AND DO NOT ADD ANY OPTIONS TO IT
\usepackage{caption} % DO NOT CHANGE THIS AND DO NOT ADD ANY OPTIONS TO IT
\usepackage{tabularx}
\usepackage{amsmath}
\usepackage{todonotes}

\usepackage{xcolor}
\newcommand{\fan}[1]{\textcolor{black}{#1}}

\title{Better Model Selection with a new Definition of Feature Importance}
\author{
    Fan~Fang\textsuperscript{\rm 1, 2},
    Carmine~Ventre\textsuperscript{\rm 1}, \\
    Lingbo~Li\textsuperscript{\rm 2},
    Leslie~Kanthan\textsuperscript{\rm 2},
    Fan~Wu\textsuperscript{\rm 2},
    Michail~Basios\textsuperscript{\rm 2} \\
}
\affiliations{
    \textsuperscript{\rm 1}King's College London, UK \\
    \textsuperscript{\rm 2}Turing Intelligence Technology Limited, UK
}

\begin{document}

\maketitle

\begin{abstract}
Feature importance aims at measuring how crucial each input feature is for model prediction. It is widely used in feature engineering, model selection and explainable artificial intelligence (XAI). 
In this paper, 
%we extend generalisation theory of feature importance of Fisher et~al~\citep{fisher2019all} which satisfies variable importance for multiple features condition. Further, 
we propose a new tree-model explanation approach for model selection. Our novel concept leverages the \emph{Coefficient of Variation} of a feature weight (measured in terms of the contribution of the feature to the prediction) to capture the dispersion of importance over samples. Extensive experimental results show that our novel feature explanation performs better than general cross validation method in model selection both in terms of time efficiency and accuracy performance. 
% Future works conclusions
%The new explanation method for tree-structure models can also be extended to research explainability of complex tree models.
\end{abstract}

\section{Introduction} 
Model selection is the task of selecting a statistical model from a set of candidate models for given data. Cross validation is arguably the most used technique used to estimate the risk of an estimator or to perform model selection~\citep{arlot2010survey}. But cross validation through data splitting %and double cross validation % carven: meaning? % fan: agree to delete this double cross validation 
provides little additional information during the evaluation of the model and, importantly, costs a long time in retraining the model~\citep{kozak2003does}. Variable importance (a.k.a., feature importance) represents the statistical significance of the impact of each variable in the data on the generated machine learning models~\citep{strobl2008conditional}. Variable importance also related to models' explanation. For example, it can be used to measure the increase of model prediction error after replacing the \fan{object features} and, in turns, breaking the relationship between the features and the real results~\citep{breiman2001random}. \fan{Tree models are divided into black tree models (hard to explain to humans) like RandomForestTree and white box model tree models (easy to explain) like DecisionTree~\citep{rudin2019stop}. Compared to other complex black box models, such as neural networks, it is relatively easy to understand (and explain) the contributions that each variable makes to the decision of tree models~\citep{molnar2019interpretable}. Furthermore, ``features" are the core hyper-parameters in all tree models, which means it is important to combine ``features" with model selection and/or model explanation in tree models.}

Related work on Explainable artificial intelligence (XAI) explains machine learning models from features' statistic and visualization like Partial Dependence Plot~\citep{friedman2001greedy}, Individual Conditional Expectation~\citep{goldstein2015peeking} and Accumulated Local Effects~\citep{apley2016visualizing}. As from above, the importance of the features is measured by the increase of the prediction error of the calculated model after arranging the features. If shuffling the values of a feature increases model error, the feature is ``important"~\citep{fisher2019all}. This method gives feature importance an interpretation that it is the increase in model error when the feature's information is destroyed~\citep{molnar2020interpretable}. 
\fan{Related research in model selection usually involves appropriate criteria, usually based on an estimate of the generalization error, such as $k$-fold cross validation~\citep{mclachlan2005analyzing}. Other complex model selection methods, like model selection using combinatorial optimization and genetic algorithms have also been proposed~\citep{bies2006genetic}.} %\todo{What is the advantage of using our method opposed to this? Are they used in practice? Is it only of academic interest?} %fan: answer this question below:
\fan{Our approach in the area is brand new in that it focuses on applying feature explanation in model selection, combining the comprehension (and more widely the explainability) of the models with their selection. We here initiate this research by looking at tree models.}

This paper aims at proposing a new tree-model %\todo{You should try and motivate the focus on tree models? Why are they important? Why do you focus on them?}  % fan: I write this to the first paragraph.
explanation approach and apply it to model selection. Specifically, we design a new explanation estimation of tree-structure models (including DecisionTree, RandomForest, ExtraTree, GradientBoostingTree and XgboostTree) considering weighted contribution of features, the performance of contribution and feature quantity (i.e., the number of the features). This novel conceptual contribution connects common feature weighting techniques with the so-called Coefficient of Variation \cite{brown2012applied}, a statistical measure defined as the ratio of the standard deviation to the mean. We use this notion to define a new pipeline for model selection. We compare our new method with $k$-fold cross validation in model selection in five standard data sets. The results show that our method captures the dispersion of importance over samples and performs better than general cross validation method in model selection considering time efficiency and accuracy performance. The highlight is that we can generally maintain (and often improve) the performance of cross validation (in terms of test accuracy) whilst reducing computation time by \emph{at least a third}.

%\todo{Related works section on model selection and perhaps XAI?}
% fan: About XAI: I have wrote related contents in second paragraph (Basically I talk about features related XAI method because I think it is enough for readers to know these XAI. I have not talked about other methods like SHAP because it needs more explanation for this method). About model selection: Add some contents in second paragraph.

\section{Related Work}
A large body of recent research has been devoted to the estimation of the performance of a machine learning model, including (i) estimating the generalization performance of models on future data; and, (ii) selecting the best performing model from a given hypothesis space
% carven: (iii) is unclear -- seems the same as (ii)
%; and (iii) selecting the best performing model and the best performing model from the hypothesis space of the algorithm
~\citep{raschka2018model}. Basic selection/evaluation methods like Resubstitution validation, Stratified resampling and Holdout validation were proposed to be effective in selecting ``good" models. With the development of machine learning, a large numbers of settings (hyperparameters) need to be specified. Hyperparameter tuning allows to find the balance between bias and variance when optimizing the performance of these models. Cross validation is a great improvement based on holdout method in evaluating hyperparameters~\citep{kohavi1995study}. Cross validation was proposed to help to select models with a better (average) generalization than just relying on the training score~\citep{schaffer1993selecting}. In particular, in $k$-fold cross validation (aka, repeated hold-out method) the data is split in $k$ chunks and each chunk is used for testing the model trained on the remaining $k-1$ folds. Models selected by this method generally get results that are less biased and less optimistic than other methods. 

As the core definition in machine learning research, ``features" refer to an multi-dimensional vector representing the (numerical) characteristics of an object. They are core to many fields of machine learning research including dimension reduction, relevance research, automation and model explanations~\citep{zheng2018feature}. In Explainable AI (XAI), features are a medium for humans to understand the machine learning models that are hard to explain (commonly known as ``Black box models")~\citep{rai2020explainable}. Feature interaction is a method to explain models by understanding whether features affect each other and to what extent they interact. Variable Interaction Networks are a tool proposed to decompose the prediction function into main effects and feature interactions and then visualize those as a network~\citep{hooker2004discovering}. Partial dependence based feature interaction is applied in measuring the feature importance by calculating the variance of the partial dependence function~\citep{greenwell2018simple}, which illustrates and explains interaction among features in machine learning models.
But feature interaction is computationally expensive, and if we do not use all of the data points, the estimate has a non-negligible variance.

Permutation feature importance (PFI) is a concept to explain models by calculating the increase of model prediction error after the feature values are permuted. The PFI method was introduced by~\citet{breiman2001random} for random forest first. Based on idea of PFI, a model-agnostic version of the feature importance was proposed (called model reliance)~\citep{fisher2019all}. The PFI method considers both the influence of the main feature effect and the interaction effects on the performance of the model. 
But this approach has obvious drawbacks. Notably, if the features are related, the ranking  of the importance of the features may be biased by unrealistic data instances. Moreover, the importance of the associated feature might be decreased by adding a correlated feature.

Recently, a new interpretability of tree-based models is proposed. \citet{lundberg2020local} proposed to improve the explanation of tree along three dimensions: polynomial time algorithm for optimal interpretation of time based on game theory, direct interpretation of local feature interactions and understanding global model structure based on the combination of many local descriptions of each prediction. This research improves interpretability of tree-based machine learning models.

\section{Explanation of Tree-Based Models}
Variable importance describes the contribution of covariates to the prediction and model accuracy~\citep{fisher2019all}. While this works well in linear models, tree models have different variable importance calculation. Linear models' variable importance and explanation use loss functions which map the value of one or more event/variables to a real number that intuitively represents the ``cost" associated with the event to optimize the original linear models. Tree-structure models have specific realization, which separates ``nodes" and ``edges". Moreover, the splitting process will continue until no further revenue can be obtained or the preset rules are met~\citep{friedman2001elements}.

The variable importance of tree-structure models is calculated by Mean Decrease in Impurity (MDI)~\citep{louppe2013understanding} of the node, and the probability of the impurity reaching the node is obtained. The node probability can be calculated by dividing the number of samples arriving at the node by the total number of samples. The higher the value, the more important the node. A normal method is to arrange the value of each feature one by one and check how it changes the performance of the model or calculates the amount of ``impurities". But experience tells us explanation through this method it is hard to understand and link decisions with insights into actual data. Alternative method is to iterate through all the splits that use this element, and measure the degree to which the variance or Gini coefficient is reduced compared to the parent node. The sum of all importance is scaled to 100. This means that each importance can be explained as part of the overall model importance~\citep{molnar2019interpretable}.

Researchers also consider using decision paths to explain tree-structure models, which consist of each decision path from the root of the tree to the leaf. Every decision path contributes to the final prediction~\citep{guidotti2018survey}. This decision function returns a value at the correct leaf of the tree, but it ignores the operational aspect of the decision tree, namely the path through the decision node and the information available there. Since each decision path is determined by features, and the decision will be added or subtracted from the value given in the parent node, the prediction can be defined as the sum of feature contributions plus "bias" covering the entire training set area. As defined in Tree-Interpreter library in Python, the prediction function can be defined as
\begin{equation}
    f(x) = c_{full} + \sum_{k=1}^{K}contrib(x,k)%\tag{3.1}
\label{eq:1}
\end{equation}
where $K$ is the number of features, $c_{full}$ is the value at the root node \fan{calculated according to information gain }%\todo{how is this value defined?} %fan: I have add explanation and this is a single value in the root of tree.
and $contrib(x,k)$ is the contribution from the $k$-th feature in the feature vector $x$. In tree-structure models, contribution of each feature depends on the rest of the feature vector, which determines the decision path of traversing the decision tree, thereby determining the protection/contribution passed along. 

We build upon \eqref{eq:1} as follows. The specific feature $f(x,k)$ can be defined
\begin{equation}
    f(x,k) = c_{full} + contrib(x,k)%\tag{3.2}
\label{eq:2}
\end{equation}
Tree-structure models use some hyper-parameters like ``depth" and ``max leaf" to control the complexity of tree models, given data set with features. To some extent, these parameters help tree models realize dimension reduction of decision rules to improve accuracy of the classifier, which is feature selection~\citep{khalid2014survey}. Meanwhile, the quantity of features affects the explanation of machine learning models. We assume that good explanation combined with characteristics of features may lead to improvement of tree-structure models performance. In other words, we must combine the performance of feature performance with complexity of features (quantity) when we apply explanation in tree-structure models. We define the weight (contribution) of feature $k$ in the model as 
\begin{equation}\nonumber
        weight(x,k) = \frac{f(x,k)}{f(x)}  %\tag{A.1}
\end{equation}
We then consider using Coefficient of Variation~\citep{brown2012applied} as explanation of model $f$ with feature $k$, and obtain
\begin{equation}\nonumber
        Explain_{cv}(f) = \frac{\sum_{k=1}^{K}(weight(x,k)-\overline{weight})^2}{K*\overline{weight}}, %\tag{A.2}
\end{equation}
where $\overline{weight}$ is the mean of weights from 1 to K, i.e., $\overline{weight} = \frac{1}{K}\sum_{k=1}^K weight(x,k)$. From the definition of $weight(x,k)$, we know that
\begin{equation}\nonumber
        \sum_{k=1}^K weight(x,k) = 1. %\tag{A.3}
\end{equation}
We then get
\begin{equation}
    Explain_{cv}(f) = \sum_{k=1}^{K}\left(\frac{f(x,k)}{f(x)}-\frac{1}{K}\right)^2. %\tag{3.3}
\label{eq:explaincv}
\end{equation}
Eq~\ref{eq:explaincv} is defined as feature explanation of tree-structure model $f$. In this equation, $\frac{f(x,k)}{f(x)}$ is the weight (contribution) of specific feature $k$ in the construction of tree-structure models; $K$ is the quantity of features used in the model. Eq~\ref{eq:explaincv} is derived from \textit{coefficient of variation} of feature weights' contribution to the entire tree model. From the theoretical analysis, Coefficient of Variation measures the dispersion of data point around the mean~\citep{brown2012applied}. When we apply the Coefficient of Variation to the weight of features, we weigh contribution of features, the performance of contribution and quantity. Our hypothesis is that the original explanation combined with the coefficient of variation has better performance in explaining models considering various features. 

\begin{figure}[tb]
\centering
\includegraphics[width=0.9\columnwidth]{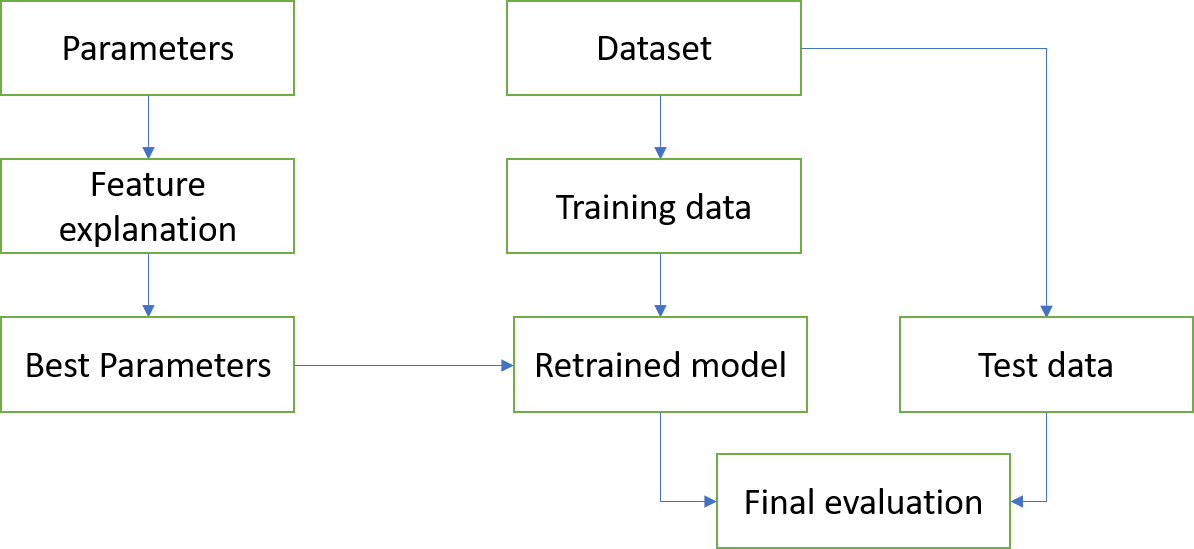}
\caption{Workflow in training data using feature explanation}
\label{fig:fig1}
\end{figure}

Considering the characteristics of feature explanation we defined, we design the workflow in model training (depicted in Figure~\ref{fig:fig1}). %\todo{Confusing, figure 1 is not about cross validation} %fan: I delete this sentence. To explain your problem: Figure 1 is the workflow contained from model selection to evaluation. When I use feature explanation method, I just use the workflow in Figure 1, when I use cross validation method, I replace ``Feature explanation" to ``cross validation" and all the same. So I have not drawn two figure in my paper.
In particular, when compared to the state of the art, we substitute cross-validation with feature explanation. Cross validation costs a lot of time because the validation mechanism needs to train data set in a loop, which is computationally expensive. More importantly, feature explanation has the function of choosing best parameters considering the training accuracy, features' contribution and feature complexity (quantity). 

\begin{figure}[b]
\centering
\includegraphics[width=0.9\columnwidth]{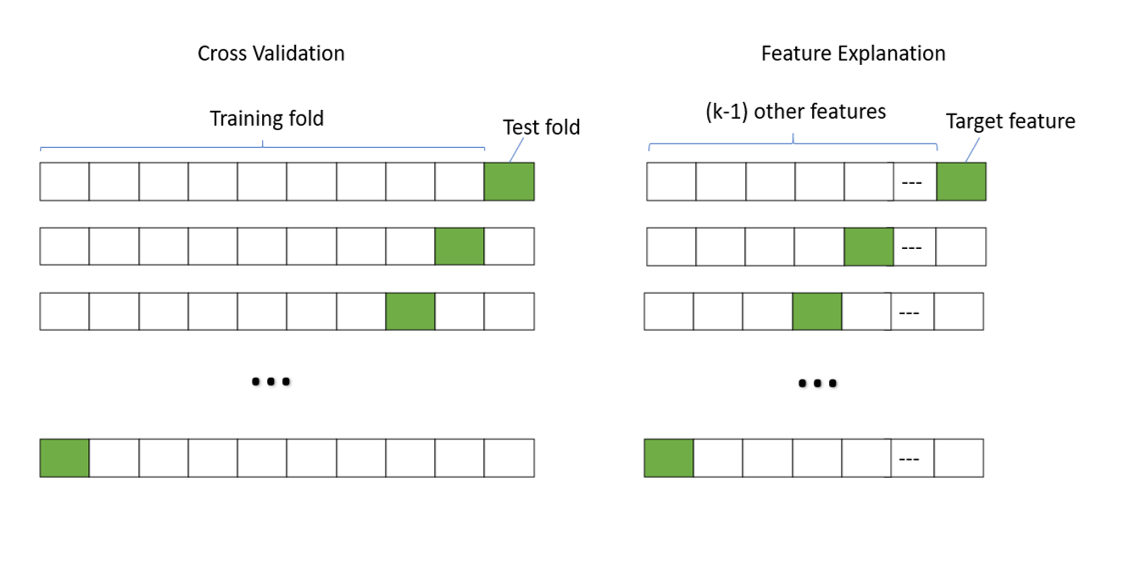}
\caption{Comparison between cross validation and feature explanation}
\label{fig:fig2}
\end{figure}

Figure~\ref{fig:fig2} describes the comparison between cross validation and feature explanation. We use $k$-fold cross validation as an example. Considering the cross validation has $k$ fold iterations and each iteration has an evaluation accuracy $E_{i}$, the final cross validation result is $E = \frac{1}{k}\sum_{i=1}^{k}E_{i}$. Each iteration needs a new training for original model. Meanwhile feature explanation splits data based on features. This method does not need a retrain of model we have trained before. We considered using feature explanation in Eq~\ref{eq:explaincv} to evaluate (even explain) models from training accuracy, features contribution and model complexity (quantity of features).

%characteristics, Num of Instances, Num of Features

\section{Experimental Study}
We evaluate the performance of our feature explanation method using tree-structure models with standard data sets. In the test process, we simulate the workflow in Figure~\ref{fig:fig1}. Our results are evaluated by test accuracy and time efficiency according to the models we selected. We considered $k$-fold as a benchmark cross validation method compared to feature explanation method.

\subsection{Experiment Subjects}
``Features" are the core hyperparameter of all tree models, which means that it is important to combine "features" with model selection and/or model interpretation in tree models. Considering characteristics of hyper-parameters in tree-structure models, we designed explanation based on feature selection for tree-structure models. Specifically, we select Decision Tree, Random Forest, Extra Trees, Gradient Boosting and XGBClassifier for simulation. Hyper-parameters used in these models are through parameter tuning~\citep{lavesson2006quantifying} with available parameters in models. To tree-structure models, important features as ``max depth", ``min sample leaf" and ``criterion" are included; ``max features" is limited by quantity of features.

\begin{table}[tb]
\centering
\caption{This table shows the property of data sets used in experimental study. The \#Classes, \#Instances, \#Features and \#Attribute present the number of classes, instances, features and attribute characteristics, respectively.}
\resizebox{.45\textwidth}{!}{
\begin{tabular}{lllll}\hline
                & \#Classes & \#Instances & \#Features & \#Attribute \\ \hline
Breast Cancer   & 2       & 286                 & 9                  & Real                      \\
Indian Diabetes & 2       & 768                 & 8                  & Integer, Real             \\
Iris            & 3       & 150                 & 4                  & Real                      \\
Bank loan       & 2       & 5000                & 10                 & Real                      \\
Wine            & 3       & 178                 & 13                 & Integer, Real             \\ \hline
\end{tabular}
}
\label{tab:datasets}
\end{table}

\subsection{Datasets}
We list details of data sets we have used in the experiment (cf. Table~\ref{tab:datasets}) for both binary and multi-object classification.
We choose some classification datasets encompassing different areas, complexities and data size. The standard data sets contains Breast Cancer Wisconsin Dataset~\citep{street1993nuclear}, Pima Indians Diabetes Database~\citep{smith1988using}, Iris Data Set~\citep{fisher1936use}, Universal Bank Loan Data Set (from Kaggle), Wine Data Set~\citep{forina1991parvus}. The dataset will be split into training dataset and test dataset (0.7/0.3) when using feature explanation as a method. When using cross validation, we will further split 20\% of training dataset for validation.

\begin{figure*}[tb]
\centering
\includegraphics[scale=0.51 ]{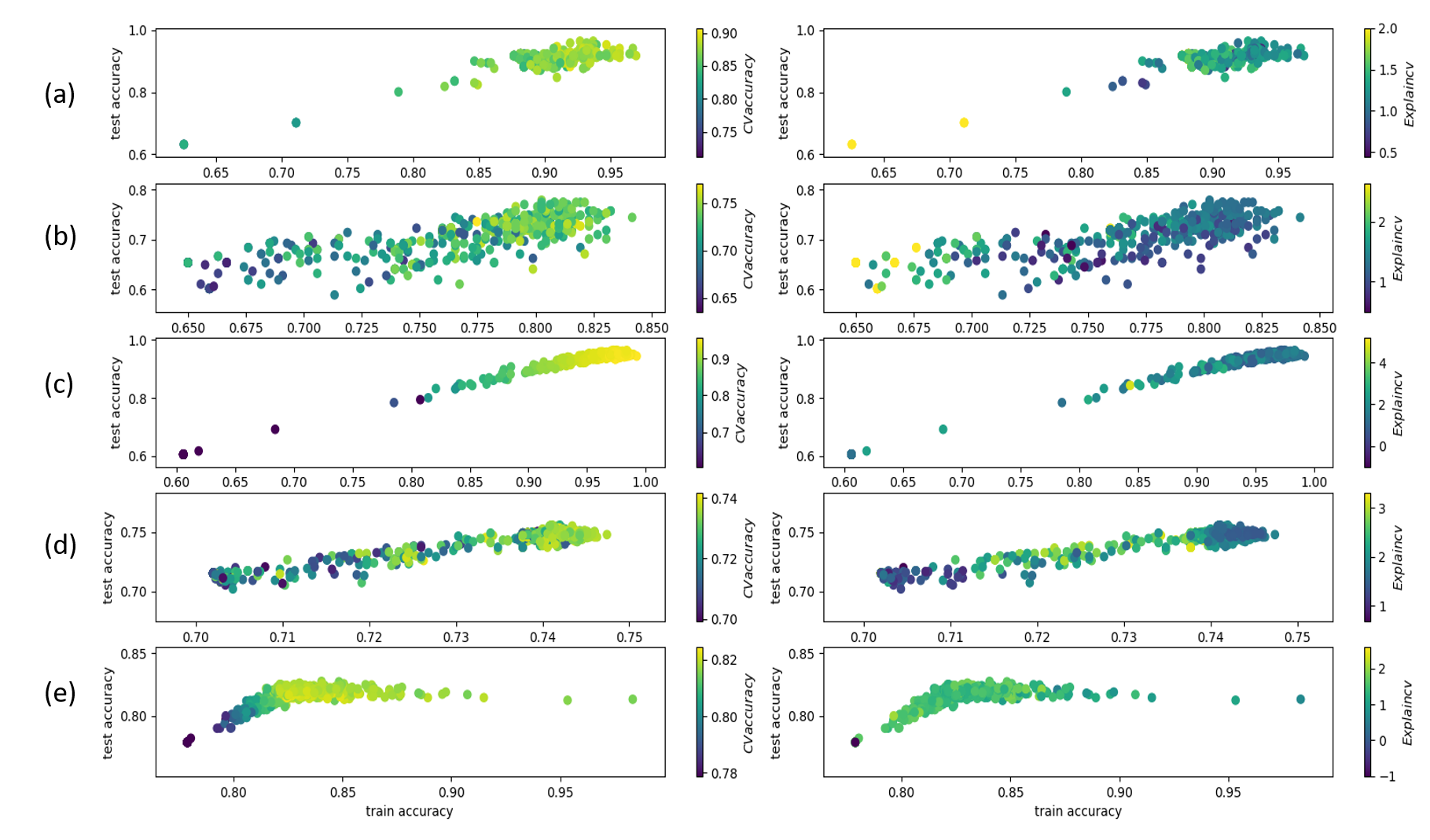}
\caption{relationship between feature explanation and accuracy matrix for the following tasks/models: (a) Breast Cancer dataset and DecisionTree, (b) Indian Diabetes dataset and DecisionTree, (c) Spam dataset~\citep{cranor1998spam} and XgboostTree, (d) Bank loan dataset and DecisionTree, (e) Credit Card dataset~\citep{yeh2009comparisons} and XgboostTree model.}
\label{fig:temp1}
\end{figure*}

\begin{table*}[tb]
\caption{This table compares performance of 10-fold cross validation method (referred to as CV) and feature explanation method (FE), using test accuracy as evaluation criteria.}
\resizebox{0.98\textwidth}{!}{
\begin{tabular}{lllllll}
\hline
                &                  & DecisionTree    & RandomForest  & ExtraTree     & GradientBoostingTree & XgboostTree   \\  \hline
Breast Cancer   & test Acc (CV/FE) & 92.78\% / 93.18\% & 96.88\%/96.30\% & 95.13\%/95.13\% & 96.68\%/96.49\%        & 95.03\%/95.78\% \\
                & exec time (s)    &  19.459 / 3.288  &  2071 / 153   &  584 / 62   &   105 / 19     &        74 / 1.897       \\
Indian Diabetes & test Acc (CV/FE) & 73.30\% / 72.58\% & 74.17\%/74.75\% & 76.76\%/75.76\% & 75.18\%/75.04\%        & 75.32\%/75.90\% \\
                & exec time (s)    &      34.668 / 5.543  &     2200 / 174   &    990 / 102      &    237 / 33   &  117 / 2.194  \\
Iris            & test Acc (CV/FE) & 100\%/100\%       & 100\%/100\%     & 100\%/100\%     & 100\%/100\%            & 100\%/100\%     \\
                & exec time (s)    &     3.976 / 1.666    &      2419 / 133  &  664 / 51     &   467 / 51        &  77 / 0.965    \\
Bank loan       & test Acc (CV/FE) & 74.67\%/74.93\%   & 73.82\%/73.78\% & 75.26\%/75.26\% & 71.56\%/71.56\%        & 73.86\%/73.47\% \\
                & exec time (s)    &   63.922 / 10.183   &     4393 / 959    &   1826 / 227  &   606 / 103  &    324 / 6.312    \\
Wine            & test Acc (CV/FE) & 93.21\%/93.83\%   & 98.76\%/99.38\% & 98.76\%/100\%   & 93.82\%/92.59\%        & 98.48\%/98.41\% \\
                & exec time (s)    &      13.793 / 5.656  &      2334 / 142 &    558 / 56      &   660 / 76   &  114 / 1.268    \\ \hline        
\end{tabular}
}
\label{tbl:performance}
\end{table*}

\subsection{Evaluation Criteria}
In this research, we focus on using feature explanation in model selection. When we compare performance of cross validation based selection and model selection by feature explanation, time efficiency and accuracy (ACC) performance are two factors we consider. %Cross validation costs much time on retraining
Similar to normal model selection process, after we trained the tree models using parameter tuning with training datasets, we will use test datasets to evaluate the accuracy performance of selecting models according to accuracy matrix. 

\begin{figure}[tb]
\centering
\includegraphics[scale=0.28]{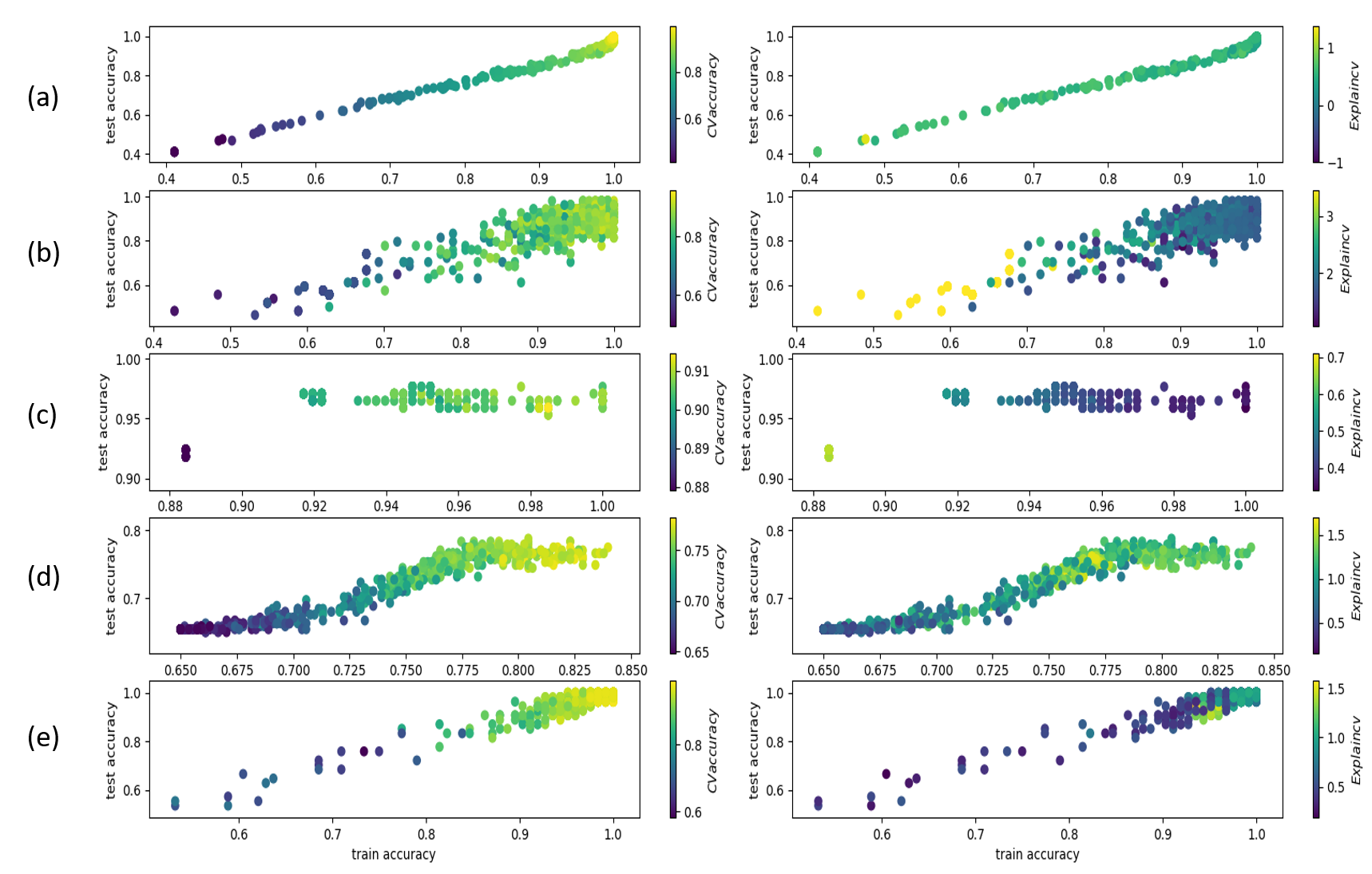}
\caption{relationship between feature explanation and accuracy matrix for the following tasks/models: (a) Avila dataset (from UCI) and XgboostTree, (b) Wine dataset and DecisionTree, (c) Breast Cancer dataset and RandomForest, (d) Indian Diabetes dataset and DecisionTree, (e) Wine dataset and Extratree model.}
\label{fig:temp2}
\end{figure}

\begin{figure}[tb]
\centering
\includegraphics[scale=0.28]{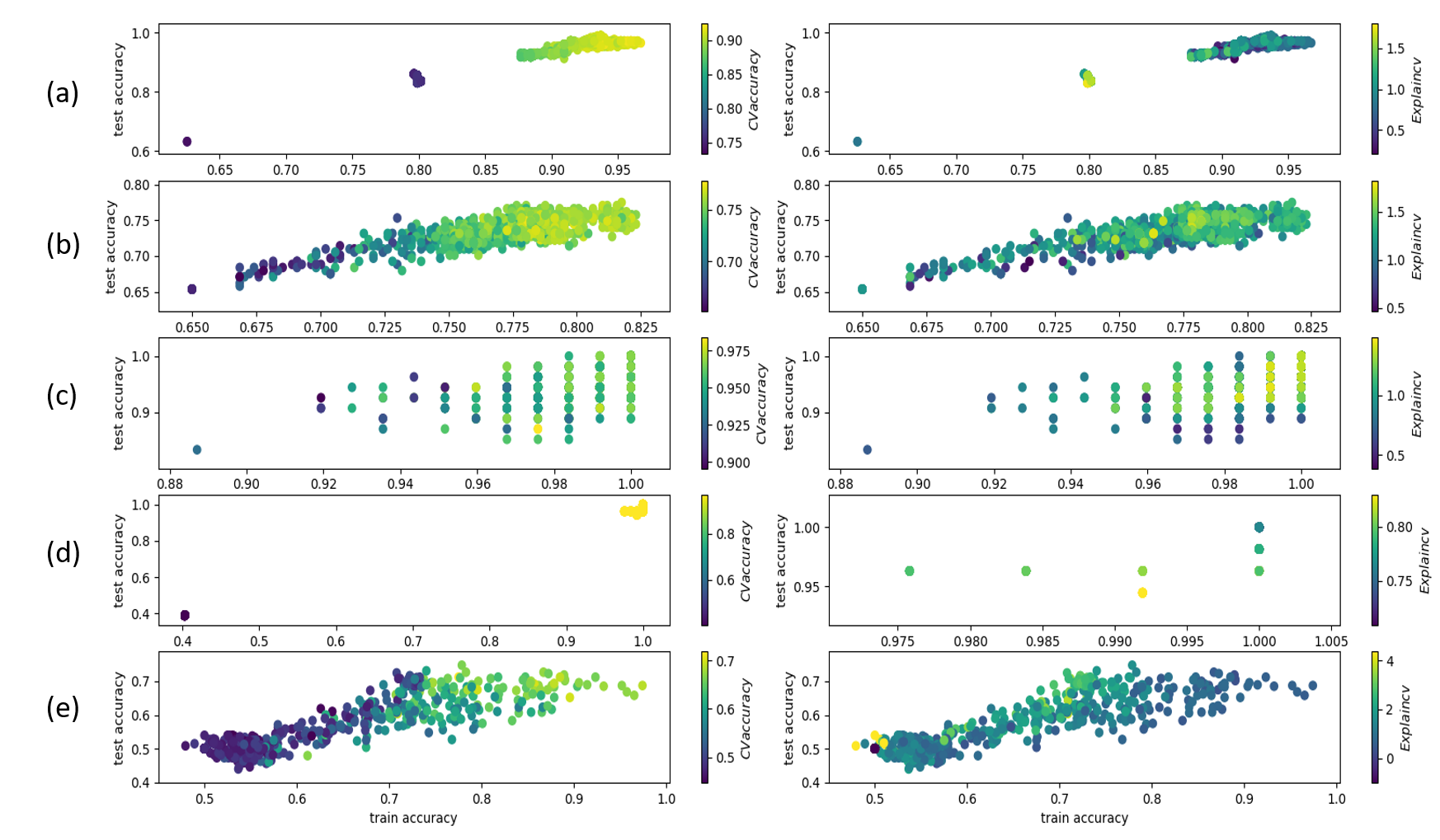}
\caption{relationship between feature explanation and accuracy matrix for the following tasks/models: (a) Breast Cancer dataset and GradientBoostingTree, (b) Indian Diabetes dataset and GradientBoostingTree, (c) Wine dataset and GradientBoostingTree, (d) Wine dataset and RandomForest, (e) Gametes1 (from OpenML) dataset and XgboostTree model.}
\label{fig:temp3}
\end{figure}

\subsection{Research Questions}
To evaluate our method and compare it to cross validation in model selection, we explore the following research questions (RQs, for short):

\begin{description}
\item[RQ1.] What is the relationship between feature explanation and accuracy matrix in model selection? How can we apply feature explanation method to select tree models?

We know that feature explanation evaluates tree models from training accuracy, features contribution and model complexity. We need to experimentally evaluate what is the relationship between explanation and (test) accuracy. Depending on the empirical findings about the relationship, we could apply feature explanation method in tree models selection.
\item[RQ2.] What is the effect to apply our feature explanation method in model selection compared to $k$-fold cross validation?

Applying the results from RQ1 about selecting tree models using feature explanation, we need to experimentally compare the performance of our method with $k$-fold cross validation. The evaluation will look at accuracy performance and computational cost (time).
\end{description}

\section{Experimental Results}
In this section, we present the results of the experimental study, and interpret the research questions sequentially and separately to explain why the proposed approach is better than traditional cross-validation in model selection. 

\subsection{RQ1: Relationship between feature explanations and accuracy matrix}
As discussed above, feature explanation explain tree models from training accuracy, features contribution and model complexity. When we apply feature explanation in model selection, we experiment the relationship between feature explanations and accuracy matrix using classification datasets. We have listed five of the results of relationship between feature explanations and accuracy matrix in Figure~\ref{fig:temp1} from experiments of Decision tree model and Xgboost tree model. Each point in the relationship figure refers to a model evaluation based on hyperparameter tuning. %In each experiment we present two figures, plotting the relationship between test accuracy and $Explain_{cv}$ we defined in Eq~\ref{eq:explaincv} (use train accuracy and cross validation accuracy as control group). 
\fan{In each experiment we present two figures, plotting the relationship between train accuracy and test accuracy. The experiment uses cross validation accuracy (CVaccuracy in figure) and $Explain_{cv}$ value (we defined in Eq~\ref{eq:explaincv}) as control group.}
When we perform model selection, the results of cross-validation accuracy is the core evaluation to select ``good" model after training the model under most cases. The \fan{left} part of this figure shows that cross validation cannot \fan{always} select models with higher test accuracy. The \fan{right} part of figure shows that when the value of feature explanation is smaller, the model is likely to achieve a higher test accuracy. It is worth noting that the distribution of models' test accuracy under cross validation and feature explanation is very similar. From Figure~\ref{fig:temp1}, cases (a) and (b) {reflect cases in which it is hard to select models with best test accuracy by cross validation or feature explanation method}, as shown by the  sparse plots in test accuracy. The other three experiments -- (c), (d), (e) -- perform effectively. When the training accuracy is high and applying feature explanation method, smaller feature explanation values can lead to higher test accuracy. From the point distribution level, cross validation and feature explanation appear to aggregate when selecting models with high test accuracy, which might because cross validation and feature explanation have similar learning effects of features' characteristics.
Figure~\ref{fig:temp2} and Figure~\ref{fig:temp3} are provided for completeness; the plots show more results for the relationship between features and accuracy matrix. In most cases, we find that low values of feature explanation lead to higher test accuracy (which means ``better" models have been found). Some cases do not have clear distinction between accuracy of cross validation and feature explanation (e.g., cases (c) and (d) in Figure~\ref{fig:temp3}), which might be caused by model overfitting and datasets that are too simple for the task at hand.

The results show that compared to cross validation, in the majority of cases, the value of feature explanation in tree models show an inverse relationship with test accuracy. When we have a high test accuracy, feature explanation method and normal cross validation method have dense distribution. This means, feature explanation allows to find models with high test accuracy when the model's feature explanation value is low.

\subsection{RQ2: Feature Explanation in Model Selection}
To evaluate the performance of feature explanation in model selection compared to cross validation method, we simulate the basic model selection on five datasets (cf. Table~\ref{tbl:performance}). ``Test Acc" means the average test accuracy from the best three models selected from cross validation method (CV) and feature explanation method (FE). The results show that feature explanation method could select same or better models than cross validation method. In general, half of experiments show that FE achieve higher test accuracy of best three selected models than cross validation methods. In some cases (for example cases in GradientBoostingTree), FE achieves slightly worse test accuracy (no more than 1\%). ``Exec time" refers to the execution time when applying CV and FE method in model selection respectively {(including the whole process of model selection, cf. Figure~\ref{fig:fig1})}. The results show that FE method saves \emph{at least 300\%} of execution time when selecting models in tree-structure models. Time efficiency of feature explanation is because the comparison of mechanism between CV and FE (cf. Figure~\ref{fig:fig2}). In model selection, the program needs to retrain and evaluate the model up to ten times (in 10-fold cross validation), which costs too much time. In feature explanation method instead, models that have been trained will be analysed by features contribution according to models' formation and features' contribution.

The results show that feature explanation performs at least as well as general cross validation method ($k$-fold) in tree-structure model selection while feature explanation method has a notable advantages in time efficiency. It means that we can safely replace general cross validation with feature explanation method in tree-structure model selection in the vast majority of cases.

\section{Threats to Validity}
The feature explanation method in this research is based on a novel notion in tree models (containing training accuracy, features contribution and model complexity). So this method is very sensitive to input features. Consider the noise of features (we consider the ``noise" as repeated or inappropriate features in datasets); selecting bias of features in datasets might impact the practical efficacy of our method. For example, in cases in which there is no benchmark feature collections (like financial market price prediction), the same model works better on datasets including more reasonable features than datasets with unreasonable ones. It means, the application of our method might take more time on feature selection in the data processing phase.

Another threat we must point out is about the settings of hyperparemeters in models. In our experimental study, hyperparemeters are selected from a reasonable search space. But we cannot cover all possible hyperparemeters in the experiments. In this research, we select most hyperparemeters that work for classification purpose.

The stochastic characteristics of model evaluation is also a threat. We are trying to mitigate this threat by running 20 times each of the experiments and choosing 5 different tree models to increase the diversity domain of all aspects.

\section{Conclusions}
% Con:  Features will interference the final results -> noise
In this paper, we propose a new tree-model explanation approach for the model selection. According to our experiments, the results show that, in the vast majority of the cases, feature explanation allows to select models with high test accuracy when the model’s feature explanation value is low. Compared to general cross validation method ($k$-fold), our method performs better or similar in model selection performance while our method has a large efficiency improvement in model selection execution time. At the same time, we noticed that feature explanation based model selection would be impacted by features selection (eliminate noise generated by useless features). %
%We provide for completeness, more results -- akin the ones in Figure \ref{fig:appen1} -- in the appendix of this paper (cf. Figures \ref{fig:appen2} and \ref{fig:appen3}) which confirm that feature explanation is a better model selection technique than cross validation. 

Possible future research directions might be the optimization of this explanation method and its application to other machine learning models.

\bibliography{Formatting-Instructions-LaTeX-2021}

% \appendix

% \section{More experimental findings}
% We here show some more experimental results, which confirm the findings and patterns discussed in the main body of the paper. 

% \begin{figure*}[!t]
% \centering
% \includegraphics[scale=0.55]{appendix1.png}
% \caption{Correlation between feature explanation and accuracy matrix for the following tasks/models: (a) Avila dataset (from UCI) and XgboostTree, (b) Wine dataset and DecisionTree, (c) Breast Cancer dataset and RandomForest, (d) Indian Diabetes dataset and DecisionTree, (e) Wine dataset and Extratree model.}
% \label{fig:appen2}
% \end{figure*}

% \begin{figure*}[!t]
% \centering
% \includegraphics[scale=0.55]{appendix2.png}
% \caption{Correlation between feature explanation and accuracy matrix for the following tasks/models: (a) Breast Cancer dataset and GradientBoostingTree, (b) Indian Diabetes dataset and GradientBoostingTree, (c) Wine dataset and GradientBoostingTree, (d) Wine dataset and RandomForest, (e) Gametes1 (from OpenML) dataset and XgboostTree model.}
% \label{fig:appen3}
% \end{figure*}

% \begin{figure}[!t]
% \centering
% \includegraphics[scale=0.31]{temp1.png}
% \caption{Correlation between feature explanation and accuracy matrix for the following tasks/models: (a) Avila dataset (from UCI) and XgboostTree, (b) Wine dataset and DecisionTree, (c) Breast Cancer dataset and RandomForest, (d) Indian Diabetes dataset and DecisionTree, (e) Wine dataset and Extratree model.}
% \label{fig:appen2}
% \end{figure}

\end{document}